\documentclass[conference]{IEEEtran}
\IEEEoverridecommandlockouts
\usepackage{cite}
\usepackage{amsmath,amssymb,amsfonts}
\usepackage{mathtools}
\usepackage{graphicx}
\usepackage{textcomp}
\usepackage{stmaryrd}
\usepackage{subcaption}
\usepackage{multirow}
\usepackage{array}
\usepackage{xurl}
\usepackage{xcolor}
\usepackage{hyperref}
\hypersetup{colorlinks, citecolor=green, filecolor=pink, linkcolor=blue, urlcolor=blue }
\def\BibTeX{{\rm B\kern-.05em{\sc i\kern-.025em b}\kern-.08em
    T\kern-.1667em\lower.7ex\hbox{E}\kern-.125emX}}
\usepackage[T1]{fontenc}
\usepackage[utf8]{inputenc}
\usepackage{authblk}
\usepackage{xurl}
\usepackage{array}
\usepackage{algorithm}
\usepackage{algpseudocode}

\algrenewcommand\algorithmicrequire{\textbf{Input:}}
\algrenewcommand\algorithmicensure{\textbf{Output:}}
\newcommand\T{\rule{0pt}{2.6ex}}       
\newcommand\B{\rule[-1.2ex]{0pt}{0pt}} 
\newcolumntype{C}[1]{>{\centering\let\newline\\\arraybackslash\hspace{0pt}}m{#1}}

\begin{document}


\title{Sustainable AIGC Workload Scheduling of Geo-Distributed Data Centers: A Multi-Agent Reinforcement Learning Approach 
}



\author{Siyue Zhang, Minrui Xu, Wei Yang Bryan Lim, and Dusit Niyato, \emph{Fellow, IEEE}
  \thanks{ Siyue~Zhang and Wei~Yang~Bryan~Lim are with Alibaba-NTU Singapore Joint Research Institute, Singapore 637335, Singapore and the School of Computer Science and Engineering, Nanyang Technological University, Singapore 639798, Singapore (e-mail: siyue001@e.ntu.edu.sg; bryan.limwy@ntu.edu.sg). Minrui~Xu and Dusit~Niyato are with the School of Computer Science and Engineering, Nanyang Technological University, Singapore 639798, Singapore (e-mail: minrui001@e.ntu.edu.sg; dniyato@ntu.edu.sg).}
}

\maketitle

\begin{abstract}

Recent breakthroughs in generative artificial intelligence have triggered a surge in demand for machine learning training, which poses significant cost burdens and environmental challenges due to its substantial energy consumption. Scheduling training jobs among geographically distributed cloud data centers unveils the opportunity to optimize the usage of computing capacity powered by inexpensive and low-carbon energy and address the issue of workload imbalance. To tackle the challenge of multi-objective scheduling, i.e., maximizing GPU utilization while reducing operational costs, we propose an algorithm based on multi-agent reinforcement learning and actor-critic methods to learn the optimal collaborative scheduling strategy through interacting with a cloud system built with real-life workload patterns, energy prices, and carbon intensities. Compared with other algorithms, our proposed method improves the system utility by up to 28.6\% attributable to higher GPU utilization, lower energy cost, and less carbon emission.

\end{abstract}


\section{Introduction}


The rapid development of machine learning (ML) algorithms and computing capacity has contributed to the widespread success of AI in the last decade. AI has demonstrated outperforming capabilities in image and object recognition, game-playing, natural language processing, and data analysis \cite{ai-index-report}. Moreover, recent breakthroughs in Generative AI have endowed the machine with extraordinary capability in creativity, which unleashes a new paradigm, AI-Generated Content (AIGC), producing high-quality content at a large scale. A tremendous demand has been triggered, e.g., OpenAI ChatGPT crossed a whopping 1 million users within a week, and a single version of the Stable Diffusion model in Hugging Face has 1.2 million downloads in a month.


However, training, fine-tuning, and inference of AIGC models require substantial computation therefore energy, negatively impacting the environment. For example, training GPT-3, a large language model with hundreds of billions of parameters, requires 1,287 MWh energy consumption and emits 552 tonnes CO$_{2}$eq \cite{BLOOM}. The Diffusion Models (DMs) often take hundreds of GPU days (e.g. 150-1,000 V100 days) to train and 5 days to generate 50k samples on a single A100 GPU \cite{LDM}. More significantly, those large pre-trained models will be extensively fine-tuned for customized downstream tasks, creating numerous energy-intensive ML training jobs and aggravating the negative environmental impact. Consequently, the sustainability of ML training becomes a critical issue \cite{green_metaverse}. One of the solutions is to schedule jobs across geo-distributed data centers for optimized operational and environmental costs. For instance, migrating jobs from a data center in Singapore, where the energy price is 301 USD/MWh and the carbon intensity is 483 gCO$_2$eq/kWh from natural gas \cite{electricitymaps}, to a data center in Ontario, Canada, where the energy price is 22 USD/MWh and the carbon intensity is 67 gCO$_2$eq/kWh from hydropower \cite{electricitymaps}, can result in substantial advantages. Moreover, more jobs can be fulfilled during peak demand by remote data centers with surplus computing capacity, especially for different time zones and holidays.


Notwithstanding the promising benefit, scheduling ML training jobs in the global cloud computing network faces several challenges. The characteristics of ML workloads differ significantly from those of traditional workloads \cite{gao2022deep}: 
\begin{itemize}
    \item Intensive and exclusive GPU usage.
    \item Gang scheduling, i.e., simultaneous GPU allocation.
    \item Sensitivity to the locality of allocated GPUs.
    \item Large public and private datasets.
    \item Sophisticated ML models.
\end{itemize}
In addition, complex heterogeneity exists in ML training jobs such as arrival time, training data, and ML model, as well as in data centers including energy efficiency, energy prices, and carbon intensity. It is an NP-hard problem to optimize task scheduling in a distributed system for multiple objectives \cite{MURAD20222309}. In this case, our goal is to maximize the utilization of computing resources while minimizing operational costs. Research efforts have been made to develop heuristic algorithms for scheduling traditional workloads across renewable energy-powered data centers, e.g., web application requests \cite{toosi2017renewableweb}, and Google’s internal workloads \cite{green_datacenter_survey}. There is a lack of research on scheduling ML training workloads among general data centers.



Therefore, this paper proposes job scheduling in the global cloud computing network to improve computation sustainability for the first time and focuses on AIGC fine-tuning jobs. With modular pre-trained models provided by emerging Model-as-a-Service (MaaS) products, private data can be preprocessed locally once using the input layers of the pre-trained model. Intermediate results will be transmitted to remote data centers for iteratively fine-tuning the hidden and output layers of the pre-trained model. Pre-trained models are widely cached in cloud data centers. It mitigates the privacy issue by maintaining the raw data locally and reduces the data transmission of datasets and pre-trained models. To tackle the difficulty in multi-objective job scheduling, we propose a scheduling algorithm leveraging the advantages of multi-agent reinforcement learning (MARL) and Soft Actor Critic (SAC) \cite{haarnoja2018soft} algorithms, namely MASAC. MARL eliminates the single point of failure in the central scheduling system and is scalable when the network grows. SAC balances policy exploitation with action exploration optimally and has the advantage of addressing complex reward structures such as delayed rewards.

In this paper, we model the system of a cloud computing network consisting of geo-distributed data centers with real-life energy price and carbon intensity datasets from \cite{electricitymaps}. An AIGC workload is synthesized based on real-life demand patterns from Alibaba GPU Cluster Trace \cite{alibaba2022mlaas} and characteristics of AIGC fine-tuning jobs. Multiple RL agents are employed to process, postpone or migrate jobs collaboratively for optimal system utility. Our contributions are summarized as follows:
\begin{itemize}
    \item Different from previous studies working on traditional CPU-based workloads, we propose to schedule AIGC workloads in the global cloud system considering special characteristics of ML training and mitigate massive data transmission and privacy challenges by leveraging MaaS. It tackles the imbalance of computation demand and low-cost and clean energy availability among geo-distributed cloud data centers.
    \item We propose a cost and carbon-aware multi-objective job scheduling algorithm MASAC, a synergy of MARL and actor-critic algorithms with advantages in scalability, reliability, and performance.
    \item We develop a simulation environment for scheduling AIGC fine-tuning jobs, quantify the benefit of migrating jobs in the cloud system and evaluate the performances of the proposed algorithm using real-world workload, energy price, and carbon intensity datasets.
\end{itemize}

The rest of the paper is organized as follows: Section \ref{system_model} describes the cloud system model. Section \ref{MARL} presents the proposed MARL-based job scheduling algorithm with an actor-critic design. Section \ref{experiment} elaborates on the simulation setup and the experiment results of the proposed algorithms. Section \ref{con} concludes the findings in this paper.

\section{System Model}\label{system_model}

\begin{figure}[t]
\vspace{-0.2cm}
\centerline{\includegraphics[width=1\linewidth]{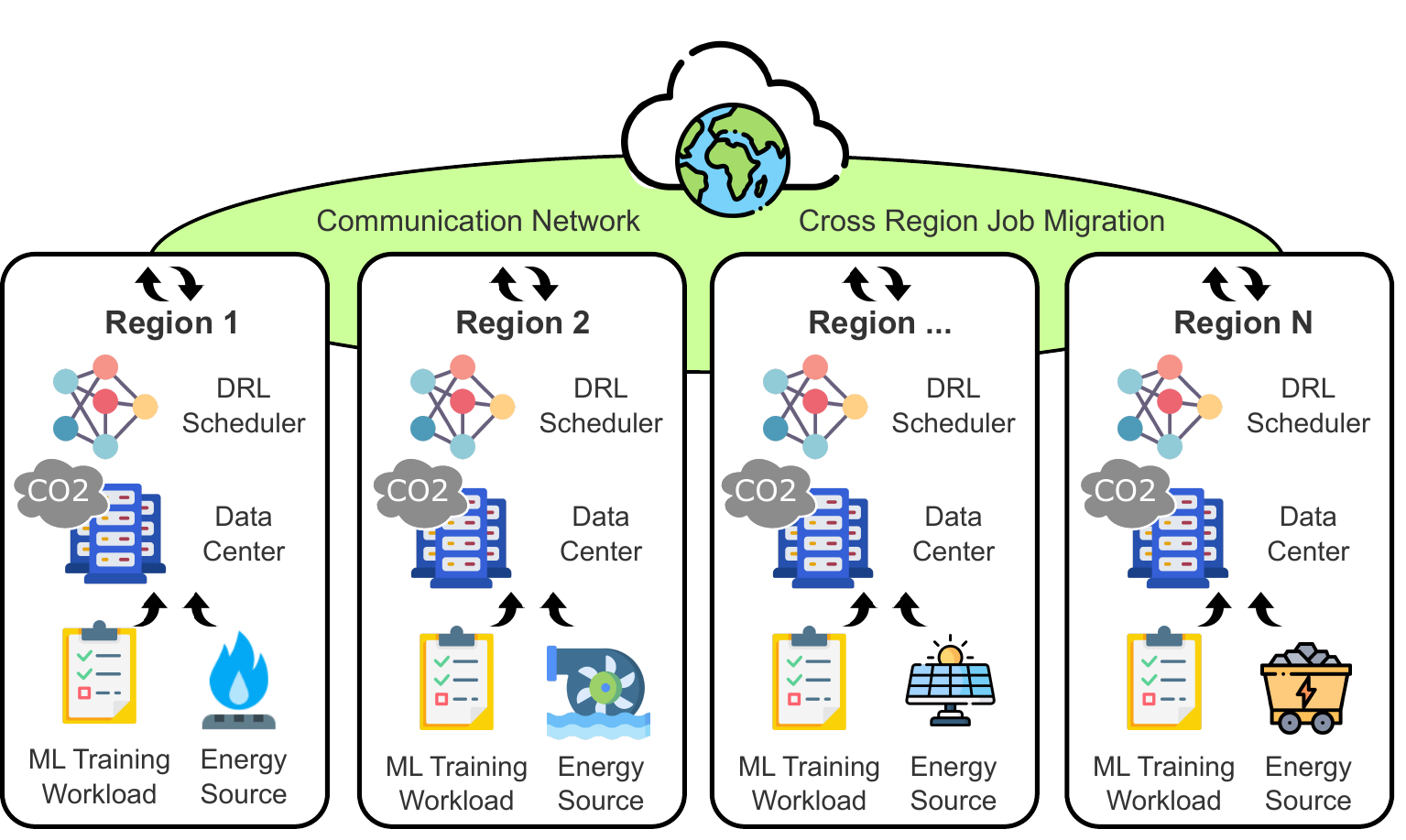}}
\caption{Cloud computing system architecture.}
\label{fig:diagram}
\end{figure}

The cloud computing system comprises three types of entities as Fig. \ref{fig:diagram}: geo-distributed data centers with diverse energy sources, a communication network relying on high-speed submarine cables, and workloads containing job requests and training datasets. We model the system as followings.


\textbf{Data center.} We consider a cloud system $\mathcal{G} = (V, E)$ consisting of $N$ data centers $V=\{v_1,\dots,v_n,\dots,v_N\}$ and edges connecting them $E = \{ e_l=(i,j) | 1 \leq i, j \leq N, i \neq j \}$. A varying amount of resources is available in each data center, including CPUs, memory, and GPUs. As GPUs are the most demanding and energy-intensive resource and CPU and memory resources are generally ample in data centers \cite{gao2022deep}, we mainly focus on optimizing GPU usage. A homogeneous GPU architecture is considered as cloud providers typically purchase GPUs in bulk. The maximum number of GPUs at data center $v_n$ is denoted by $R_n^{max}$. The data center's Power Usage Effectiveness (PUE) $\eta_n \in (1, +\infty)$ is the ratio of the total energy consumption to computing server energy consumption. The energy price at $v_n$ and time $t$ is denoted as $P_{n}(t)$, and the carbon intensity, i.e., the amount of carbon emissions per unit of energy consumed, is denoted as $I_{n}(t)$. Each data center has a queue $Q_n$ where jobs wait to be scheduled by the data center scheduler $\theta_n$.

\textbf{Workload.} The AIGC workload $J$ consisting of $M$ jobs during $T$ time slots is denoted by $J=\{j_1,\dots,j_m,\dots,j_M\}$. Each job $j_m$ is characterized by its source data center $v_{j_m}^{src} \in V$, data size $s_{j_m}$, model size $s'_{j_m}$, arrival time $t_{j_m}^{arr}$, demanded number of GPUs $r_{j_m}$, duration $d_{j_m}$, and slack $d'_{m}$. Slack is the maximum delay in starting the job that the user accepts. When a job arrives in a data center, it joins the queue and waits to be scheduled. Jobs including training datasets and model parameters can be migrated between any two data centers.


\textbf{Communication network.} The communication network is denoted by the set of edges $E$. Job migration incurs additional time delay, cost, and carbon emissions, which is essential to be considered. For transmitting data at size $s$ through the edge $e_l=(i,j) \in E$, the time delay is $\tau_{e_l}(s) = s/\xi_{i,j}$ where $\xi_{i,j}$ is the average throughput as \cite{adami2013vmmigration}. After transmission delay, the migrated job appears in $Q_j$. The migration overhead cost, defined as $c_{e_l}(s)=s \psi_{i,j}$ where $\psi_{i,j}$ is a cost coefficient as \cite{optimal}, could be significant compared to the cost of computation for small jobs. Incurred carbon emissions are $o_{e_l}(s)=\varepsilon s (I_{i}(t)+I_{j}(t))/2$, where $\varepsilon$ is the average unit energy consumption for data transmission as \cite{transmissionenergy}.


\textbf{Scheduling.} Once there is a queuing job $j_n^k$ to be scheduled in the data center $v_n$, a decision step $k$ is required. The total number of steps, denoted as $K$, is variable depending on the workload and scheduling strategy. To denote which job is being scheduled, the job status indicator $x^k_{n,m}$ equals 1 when $j_m$ is scheduled at step $k$ and data center $v_n$, otherwise 0. Thus, $ j_n^k = \sum_{m=1}^M x^k_{n,m} j_m, \sum_{m=1}^M x^k_{n,m} \leq 1, \sum_{n=1}^N x^k_{n,m} \leq 1, j_m \in J$. Simultaneously, schedulers $\theta = \{\theta_1,\dots,\theta_n,\dots,\theta_N\}$ make scheduling actions $\theta_n^k \in \{0, 1, 2, \dots, N\}$. If $\theta_n^k = 0$, the job is postponed by one time slot. If $\theta_n^k = n$, the job starts to be executed and cannot be interrupted until finished. Otherwise $\theta_n^k = d$ and $d \notin \{0,n\}$, the job will be migrated from $v_n$ to $v_d$. After all queuing jobs have been scheduled at time slot $t$, the time moves forward by one slot. Therefore, each time slot $t$ may have multiple decision steps, which are denoted by the set $K^t$. When the job $j_n^k$ is executed, the starting time $t_{j_m}^{str}=t$ and destination center $v_{j_m}^{des}=v_n$ are determined, where $k \in K^t, \theta^k_n=n, j_m=j_n^k$. Its corresponding GPU usage is
\begin{equation}
    R_{j_m}(t)=\begin{cases}
    r_{j_m},     & \text{if } t_{j_m}^{str} \leq t < t_{j_m}^{str}+d'_{j_m}\\
    0,      & \text{otherwise.}
  \end{cases}
\end{equation}
When the $j_n^k$ is migrated, i.e., $\theta^k_n = d$ and $d \notin \{0,n\}$, it is transmitted through the edge $e_l = (n,d)$. It results in transmission overhead $c_{e_l}(s_{j_n^k}+s'_{j_n^k})$, carbon emission $o_{e_l}(s_{j_n^k}+s'_{j_n^k})$, and time delay $\tau_{e_l}(s_{j_n^k}+s'_{j_n^k})$.

\textbf{Objective.} The scheduling objective is to optimize the system utility $\mathcal{U}$, which consists of GPU profit $\mathcal{U}_1$, idle GPU cost $\mathcal{U}_2$, carbon cost $\mathcal{U}_3$, job migration cost $\mathcal{U}_4$, and result retrieval cost $\mathcal{U}_5$. The GPU profit is the revenue minus energy costs associated with its operation as
\begin{equation}\label{u1}
    \mathcal{U}_1 = \sum_{t=1}^T \sum_{n=1}^N \Bigl( \alpha - \eta_n \rho P_n(t) \Bigl) \sum_{j_m \in \Theta_{n}} R_{j_m}(t), 
\end{equation}
where $J^+=\{j_n^k | \theta_n^k=n, \forall k \in [1,K], \forall n \in [1,N]\}$ denotes the set of all successful jobs, $\Theta_n= \{ j_m | j_m \in J^+, v^{des}_{j_m} = v_n \}$ denotes the set of successful jobs which are executed in the data center $v_n$, $\rho$ is the power consumption of a unit GPU time and $\alpha$ is the unit revenue per GPU time. In addition, the idle GPU cost is
\begin{equation}\label{u2}
    \mathcal{U}_2 = \sum_{t=1}^T \sum_{n=1}^N \eta_n \beta \rho  \Bigl( R_n^{max} -\sum_{j_m \in \Theta_{n}}  R_{j_m}(t)\Bigl) P_n(t),
\end{equation}
where $\beta$ is the ratio of the GPU idle power to full-load power. The carbon emission cost for both used and idle GPUs is
\begin{equation}\label{u3}
    \mathcal{U}_3 = \sum_{t=1}^T \sum_{n=1}^N \mu \eta_n \rho \biggl((1-\beta)\sum_{j_m \in \Theta_{n}} R_{j_m}(t) +  \beta R_n^{max}\biggl) I_n(t),
\end{equation}
where $\mu$ is the carbon price. To obtain the total job migration overhead and carbon emission, overhead and carbon emission incurred at each step are aggregated as
\begin{equation}\label{u4}
    \mathcal{U}_4 = \sum_{k=1}^K \sum_{n=1}^N c_{e_l}(s_{j_n^k}+s'_{j_n^k}) + \mu o_{e_l}(s_{j_n^k}+s'_{j_n^k}),
\end{equation}
where $e_l=(n, d), d = \theta^k_n, d \notin \{0,n\}$. The total cost of retrieving training results is
\begin{equation}
    \mathcal{U}_5 = \sum_{j_m \in J'^+} c_{e_l}(s'_{j_m}) + \mu o_{e_l}(s'_{j_m}),
\end{equation}
where $J'^+$ denotes the set of successful jobs which has different source and destination data centers $J'^+=\{j_m | j_m \in J^+, v_{j_m}^{des} \neq v_{j_m}^{src} \}$, and $e_l=(v_{j_m}^{des},v_{j_m}^{src})$. 

The optimization problem with the objective to maximize the utility of the global data center system is
\begin{subequations}
\begin{align}
\operatorname*{max}_{\theta} & {\mathcal{U}}  = \mathcal{U}_1 - \mathcal{U}_2 -\mathcal{U}_3 -\mathcal{U}_4 - \mathcal{U}_5 \label{eqn:line-1}\\
s.t. \quad & t_{j_m}^{str} \leq t_{j_m}^{arr} + d'_{j_m} - D'_{j_m}, \forall j_m \in J^+, \label{eqn:line-2}\\
& \sum_{j_m \in \Theta_n} R_{j_m}(t) \leq R_n^{max}, \forall t \in [1,T], \forall n \in [1,N].\label{eqn:line-3}
\end{align}
\label{eqn:all-lines}
\end{subequations}
In this problem, Eq. (\ref{eqn:line-2}) ensures the delay caused by job migration and result retrieval for each successful job is not more than its acceptable slack. Eq. (\ref{eqn:line-3}) constrains the GPU usage in a data center not beyond its maximum number of available GPUs. 


\section{MARL-Based Job Scheduling Algorithm}\label{MARL}

\begin{figure*}[t]
\centerline{\includegraphics[width=1\linewidth]{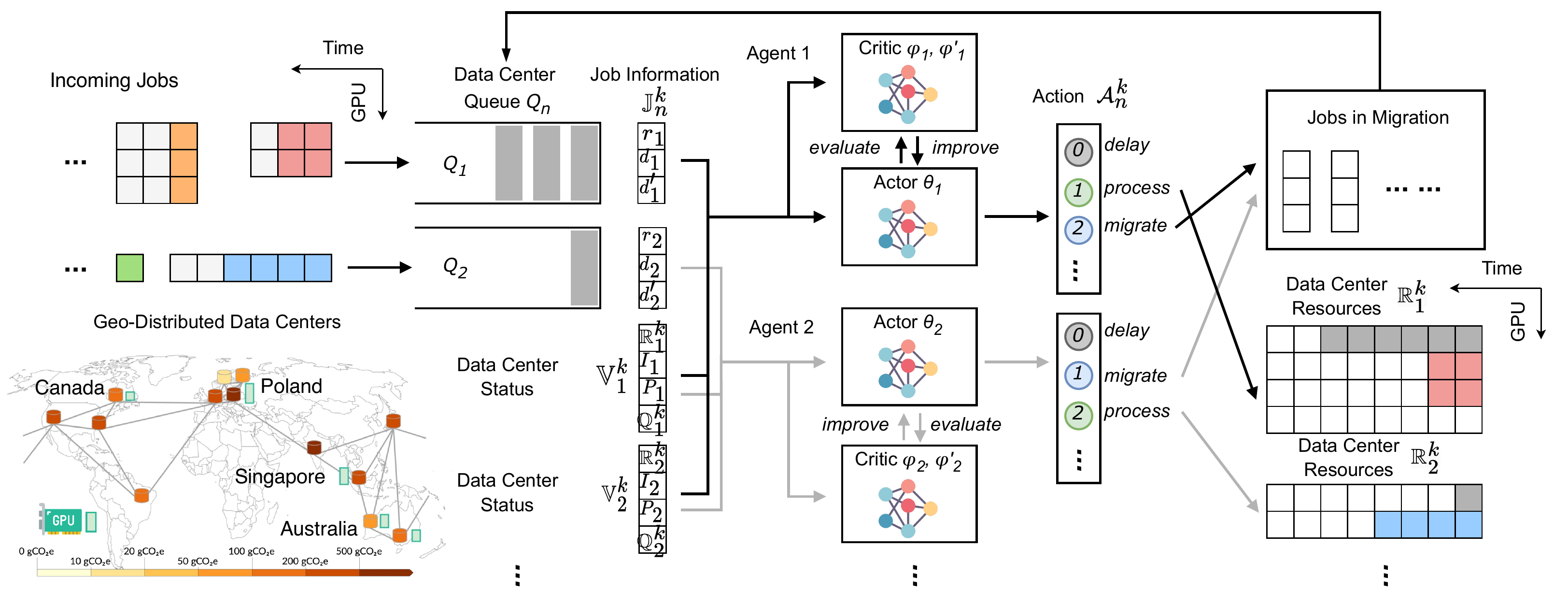}}
\caption{Flowchart of job scheduling algorithm based on MARL and actor-critic methods.}
\label{fig:system}
\end{figure*}

The job scheduling process is modeled as a Markov Decision Process (MDP) in a multi-agent environment, defined by a tuple of $\langle \mathcal{S}, \{\mathcal{A}_n\}_{n \in \mathcal{N}}, \mathcal{T}, \{\mathcal{R}_n\}_{n \in \mathcal{N}},\gamma \rangle$ as showed in Fig. \ref{fig:system}, where $\mathcal{S}$ is the state space of all agents, $\mathcal{N}=\{1,2,\dots,N\}$ is the set of agent indices, the action space is jointly defined by each agent's action space as $\mathcal{A}=\mathcal{A}_1 \times \dots \times \mathcal{A}_{N}$, $\mathcal{T}$ denotes the state transition probabilities, $\mathcal{R}_n$ represents the reward received by the agent $n$, and $\gamma$ is the factor used to discount the reward. 


\textbf{State and action.} At each step $k \in K^t$, the data center agent $n$ makes the action $\mathcal{A}_n^k = \theta_n^k$ based on its policy strategy $\pi_{\theta_n}$ and the state $S^k = \{ \langle \mathbb{J}_n^k, \mathbb{R}_n^k, \mathbb{Q}_n^k, \mathbb{V}_n^k \rangle | \forall n \in [1,N] \}$. 
$\mathbb{J}_n^k$ denotes the current job information as $\mathbb{J}_n^k = \langle R_{j_n^k}, d_{j_n^k}, d'_{j_n^k} \rangle$. $\mathbb{R}_n^k$ denotes the number of available GPUs as $\mathbb{R}_n^k = R_n^{max} - \sum_{j_m \in \Omega_n} R_{j_m}(t)$, where $\Omega_n=\{j_m | j_m \in J^+, v_{j_m}^{des}=v_n, t_{j_m}^{str} \leq t \}$ denotes the set of jobs which have been started at data center $v_n$. The total number of GPUs required for queuing jobs is denoted as $\mathbb{Q}^k_n = \sum_{j_m \in Q^k_n}{r_{j_m}}$. $\mathbb{V}^k_n$ denotes data center status including energy price and carbon intensity $\mathbb{V}_n^k = \langle P_n(t), I_n(t) \rangle$. 




\textbf{Reward.} As the objective of agents is to optimize the overall utility, the same reward is given to every agent after each step, which is defined as the system utility increment (or reduction), i.e., $\mathcal{R}^k_n = \mathcal{U}^{k+1} - \mathcal{U}^{k}, \forall n \in [1,N]$. When calculating the system utility $\mathcal{U}^k$, the time window until the step $k$ is $T=t, k \in K^t$, which is used for aggregating the GPU profits, idle power costs, and the carbon emission costs in Eqs. (\ref{u1})-(\ref{u3}). The system utility comprises both positive revenues from GPU usage and negative penalties from energy consumption, carbon emission, and job migration. Therefore, agents are rewarded to balance multiple objectives including maximizing GPU utilization, prioritizing low carbon and cost energy usage, and avoiding unnecessary data transmission.



\textbf{Policy and training.} Each agent acts based on its policy $\pi_{\theta_n}$, which is a probability distribution of discrete actions conditioned on the state. The policy is iteratively updated throughout the training, with the agent collecting trajectories of states, actions, and rewards from the environment. The actor-critic design is adopted to better estimate the long-term reward, guide the policy towards optimal actions, and address complex reward structures such as delayed rewards. Algorithm \ref{alg:cap} demonstrates the training process of MASAC algorithm. The SAC algorithm optimizes a stochastic policy in an off-policy approach and balances exploration and exploitation through entropy regularization. Besides, two critic networks $\mathcal{Q}_{\varphi}$ and $\mathcal{Q}_{\varphi'}$ are employed to stabilize the learning process. A maximum episode length $K_{max}$ is set to prevent the algorithm biased towards longer trajectories due to the entropy regularization term in the policy update.




\begin{algorithm}[ht]
\caption{MASAC Training Algorithm}\label{alg:cap}
\begin{algorithmic}[1]
\Require Job set $J$, energy price $P$, carbon intensity $I$.
\Ensure Optimized parameters $\varphi_n$, $\varphi'_n$, $\theta_n$.
\For{Agent $n = 1,2,\dots, N$}
\State Initialize empty replay buffer $\mathcal{D}_n$;
\State Initialize actor network $\pi_{\theta_n}$, first critic network $\mathcal{Q}_{\varphi_n}$, second critic network $\mathcal{Q}_{\varphi'_n}$;
\State Initialize first target network $\mathcal{Q}_{{\bar{\varphi}_n}}$, second target network $\mathcal{Q}_{{\bar{\varphi}'_n}}$;
\EndFor
\For{Epoch $e = 1,2,\dots, E$}
\For{Step $k = 1,2,\dots, K \leq K_{max}$}

\State Each agent observes state $\mathcal{S}^k$;
\State Each agent executes action $\mathcal{A}_n^k \sim \pi_{\theta_n}(\mathcal{A}_n^k, \mathcal{S}^k)$;
\State Observe system utility increment $\Delta \mathcal{U}^k$;
\State Each agent receives reward $\mathcal{R}_n^k=\Delta \mathcal{U}^k$;
\State Each agent observes next state $\mathcal{S}^{k+1}$;

\State Each agent stores $\langle \mathcal{S}^{k}, \mathcal{A}_n^k, \mathcal{R}_n^k, \mathcal{S}^{k+1} \rangle$ in replay buffer $\mathcal{D}_n$;
\EndFor

\For{Update step $g = 1,2,\dots, G$}
\State Each agent randomly samples a batch of transitions $\langle \mathcal{S}^{k}, \mathcal{A}_n^k, \mathcal{R}_n^k, \mathcal{S}^{k+1} \rangle \sim \mathcal{D}_n$;
\State Each agent updates Q-function weights $\varphi_n$, $\varphi'_n$;
\State Each agent updates policy weights $\theta_n$;
\State Each agent updates target weights $\bar{\varphi}_n$, $\bar{\varphi}'_n$.
\EndFor
\EndFor

\end{algorithmic}
\end{algorithm}

\section{Experimental Result and Analysis}\label{experiment}

\subsection{Simulation Setup}

\begin{figure}[t]
\begin{subfigure}{\linewidth}
  \centering
  \includegraphics[width=\linewidth]{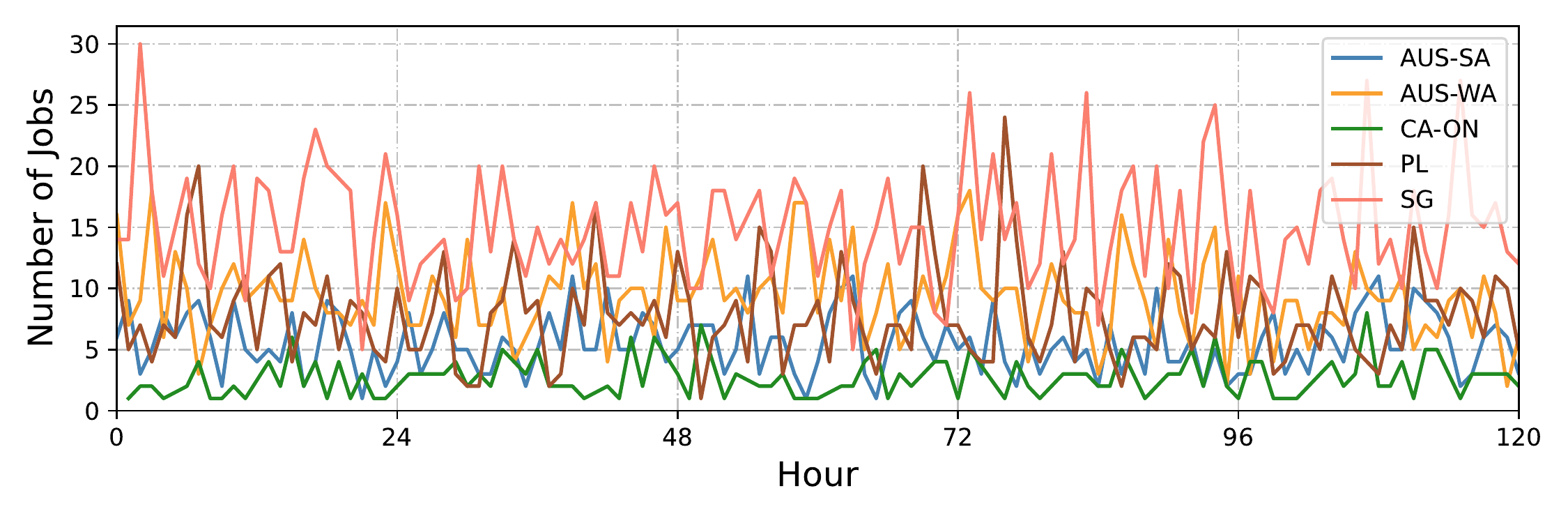}
  \vspace{-20pt}
  \subcaption{Hourly number of job arrivals.}
  \label{subfig:load}
\end{subfigure}

\begin{subfigure}{\linewidth}
  \centering
  \includegraphics[width=\linewidth]{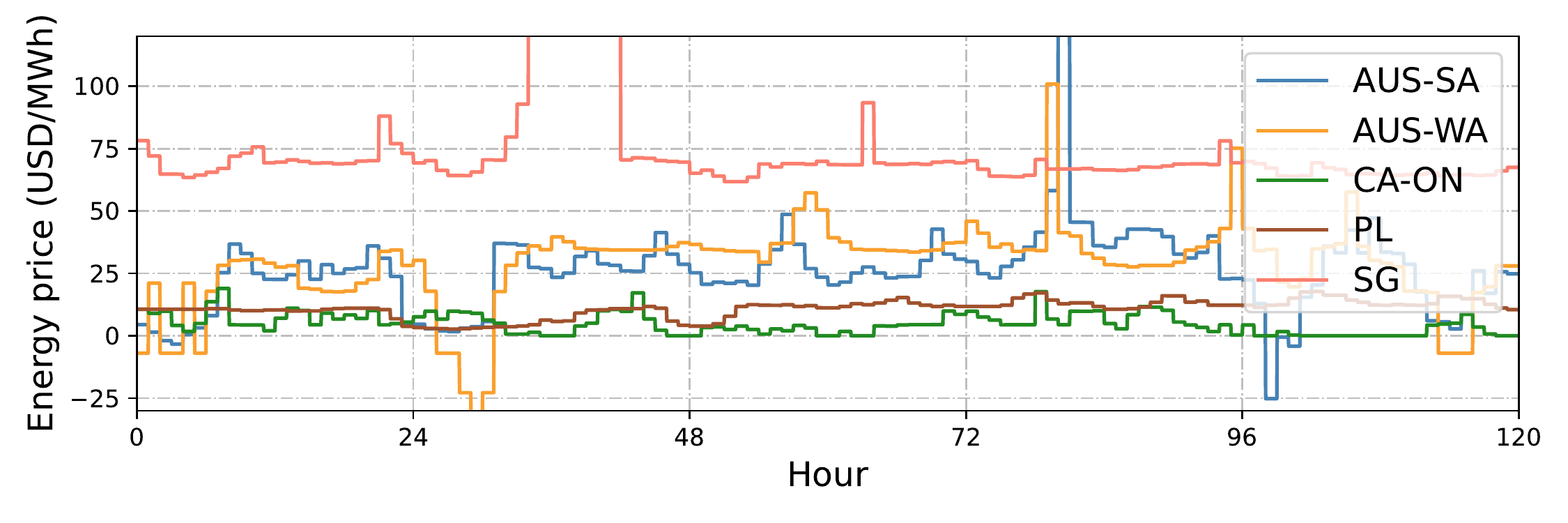}
  \vspace{-20pt}
  \subcaption{Hourly energy price (USD/MWh).}
  \label{subfig:energy}
\end{subfigure}

\begin{subfigure}{\linewidth}
  \centering
  \includegraphics[width=\linewidth]{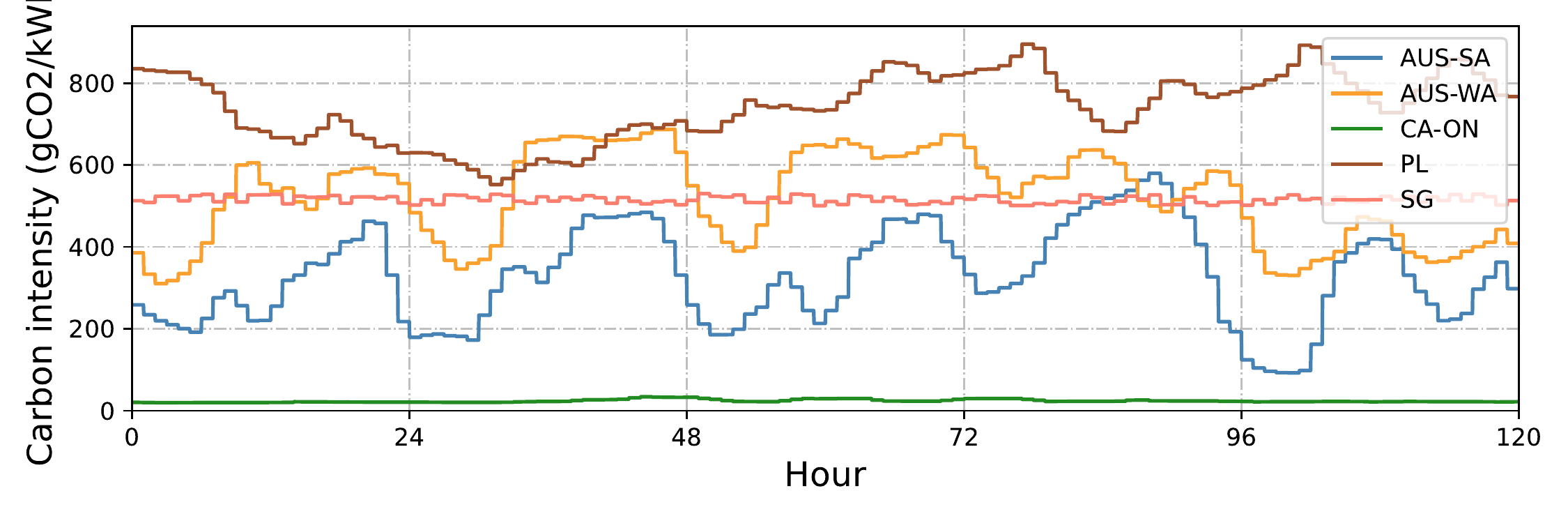}
  \vspace{-20pt}
  \subcaption{Hourly carbon intensity (gCO$_2$/kWh).}
  \label{subfig:carbon}
\end{subfigure}
\caption{Status of cloud data centers in five days (AUS-SA, AUS-WA, CA-ON, PL, and SG).}
\label{fig:all}
\end{figure}

\textbf{Simulation environment.} A simulator has been developed in Python for scheduling jobs among globally distributed data centers in discrete time steps of 1 minute. We consider 5 interconnected data centers as specified in Table \ref{table:sim_param}: South Australia (AUS-SA), West Australia (AUS-WA), Canada (CA-ON), Poland (PL), and Singapore (SG). The workload is synthesized based on the DL training job submission pattern of Alibaba GPU Cluster Trace in 2020 \cite{alibaba2022mlaas}. As the job details such as training data and model sizes are not available in the trace, we generate job characteristics based on three types of AIGC fine-tuning tasks (i.e., image generation \cite{zhao2020leveraging}, text generation \cite{mask}, and text-to-image generation \cite{pokemon}). The slack is assumed to be proportional to the job duration on average. Hourly energy price and carbon intensity series for these five locations are obtained from Electricity Maps \cite{electricitymaps}. Five days of job submission, energy price, and carbon intensity are illustrated in Fig. \ref{fig:all}, showing remarkable differences among locations. Moreover, the amount of jobs is scaled by coefficients $\delta_n$ to different levels, where SG and PL frequently experience elevated demand and insufficient GPUs.



\begin{table}[t]
    \centering
    \caption{Simulation and training parameter settings.}
    \label{table:sim_param}
    \begin{tabular}{C{3.5cm}|C{4.5cm}}
        \hline
        \textbf{Parameter} \T\B & \textbf{Value}  \T\B \\
        \hline
        Hourly GPU Revenue $\alpha$ \T & 0.05 USD \T\\
        Carbon Price $\mu$ & 100 USD/ton CO$_2$ \\
        GPU Idle Power Ratio $\beta$ & 10\% \\
        Acceptable Job Delay $d'/d$ & 40\% \\
        data center Locations $v_n$ & [AUS-SA, AUS-WA, CA-ON, PL, SG] \\
        GPU Numbers $R_n$ & [100, 110, 80, 130, 120] \\
        Datacenter PUEs $\eta_n$ & [1.3, 1.2, 1.1, 1.4, 1.2] \\
        Job Amount Coefficients $\delta_n$ \B & [0.01, 0.02, 0.01, 0.025, 0.03] \B\\
        \hline
    \end{tabular}
    \begin{tabular}{C{3.5cm}|C{4.5cm}}
        \hline
        Max Step Number $K_{max}$ & 12,000 \T\\
        Learning Rates & actor: 1e-5, critic: 5e-4 \\
        Network Update Step $G$ & 1,000 \\
        Batch Size & 256 \\
        Reward Discount $\gamma$ & 0.99 \\
        Network Hidden Layers & [256, 256] \B\\
        \hline
    \end{tabular}
\end{table}

\textbf{Training settings.} 5 RL agents are trained on 10 independent environments in parallel. Each environment has a workload of around 1,500 training jobs spanning 2 days (i.e., 2,880 mins), which are randomly selected from different months. Training parameters have been also listed in Table \ref{table:sim_param}. To monitor the policy improvement, the updated policy is validated by another environment at the end of each training epoch. After 100 epochs of training, the trained policy is tested in 10 new environments for performance evaluation without the step limit until all jobs are finished or overdue.


\textbf{Baseline scenarios.} Several cloud computing scenarios are investigated in the study, including:

\begin{itemize}
    \item Local Computing (S0): It is the real-life scenario where AIGC fine-tuning jobs are restricted within the service region selected by users and processed on the First-Come-First-Served (FCFS) basis.
    \item Greedy Migration: Jobs are primarily served by local data centers. When lacking GPUs, jobs are migrated to either the data center with the lowest energy price (S1: Price Greedy) or carbon intensity (S2: Carbon Greedy).
    \item MADQN (S3): Jobs are scheduled in the MARL approach trained by the Deep Q Network algorithm \cite{multi}.
    \item MASAC (S4): Our proposed algorithm.
\end{itemize}

\subsection{Performance Evaluation}

\begin{figure}[t]
\centering
%
\begin{subfigure}{0.8\linewidth}
  \centering
  \includegraphics[width=\linewidth]{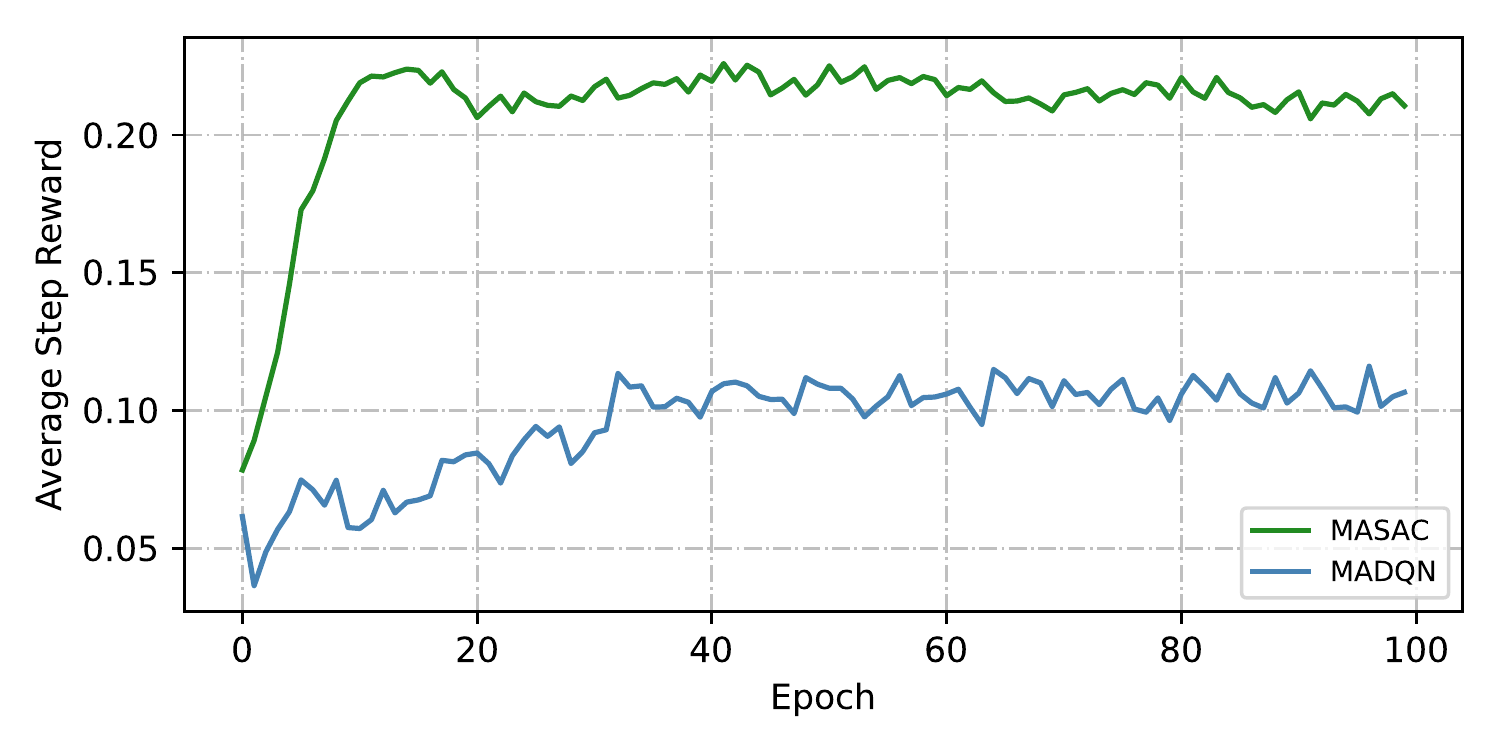}
    \vspace{-20pt}
  \subcaption{Average step reward.}
  \label{subfig:step_rew}
\end{subfigure}
\newline
\begin{subfigure}{0.75\linewidth}
  \centering
  \includegraphics[width=\linewidth]{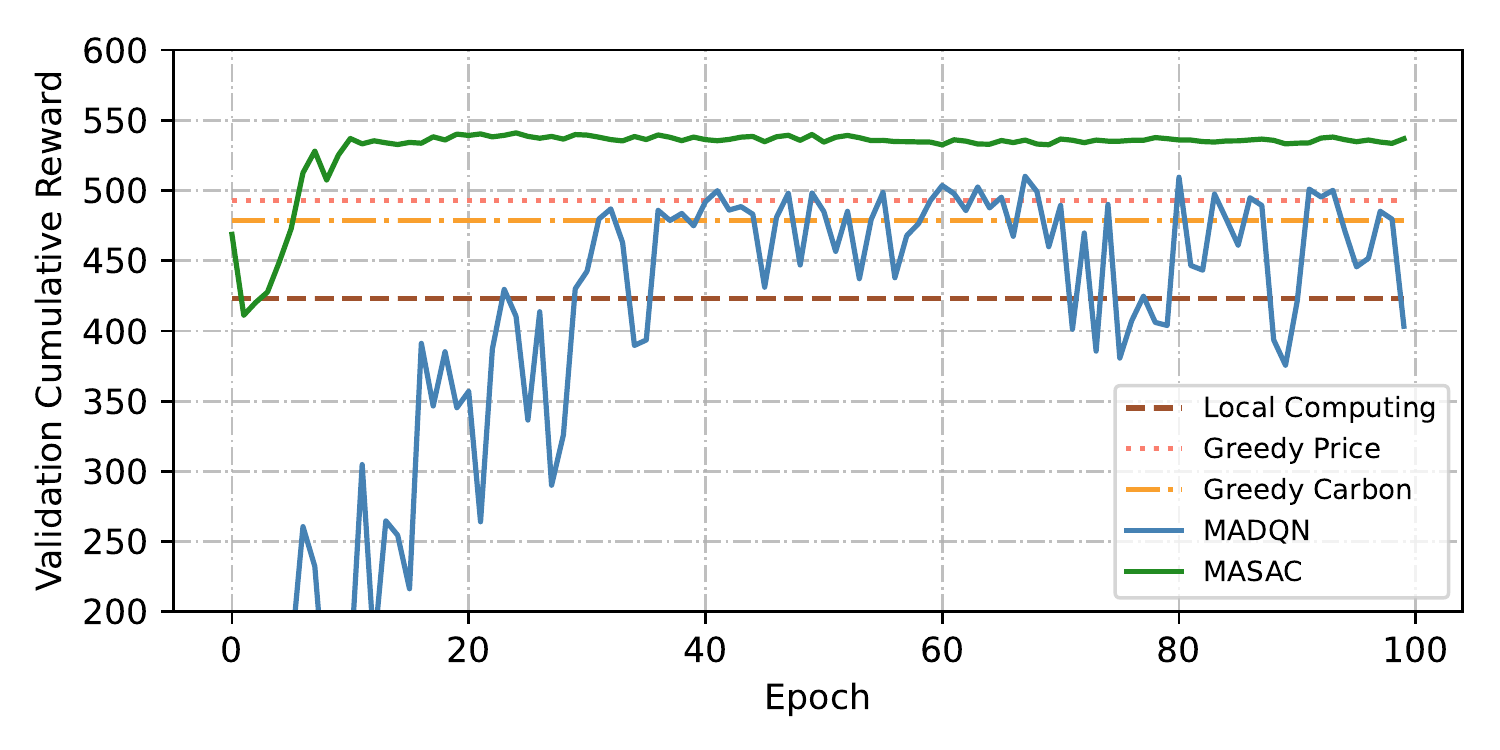}
    \vspace{-20pt}
  \subcaption{Validation reward.}
  \label{subfig:val_rew}
\end{subfigure}
    \vspace{-5pt}
\caption{Training performance of different scenarios.}
\label{fig:train_performance}
\end{figure}

\begin{figure}[t]
\centering
\begin{subfigure}{1\linewidth}
  \centering
  \includegraphics[width=\linewidth]{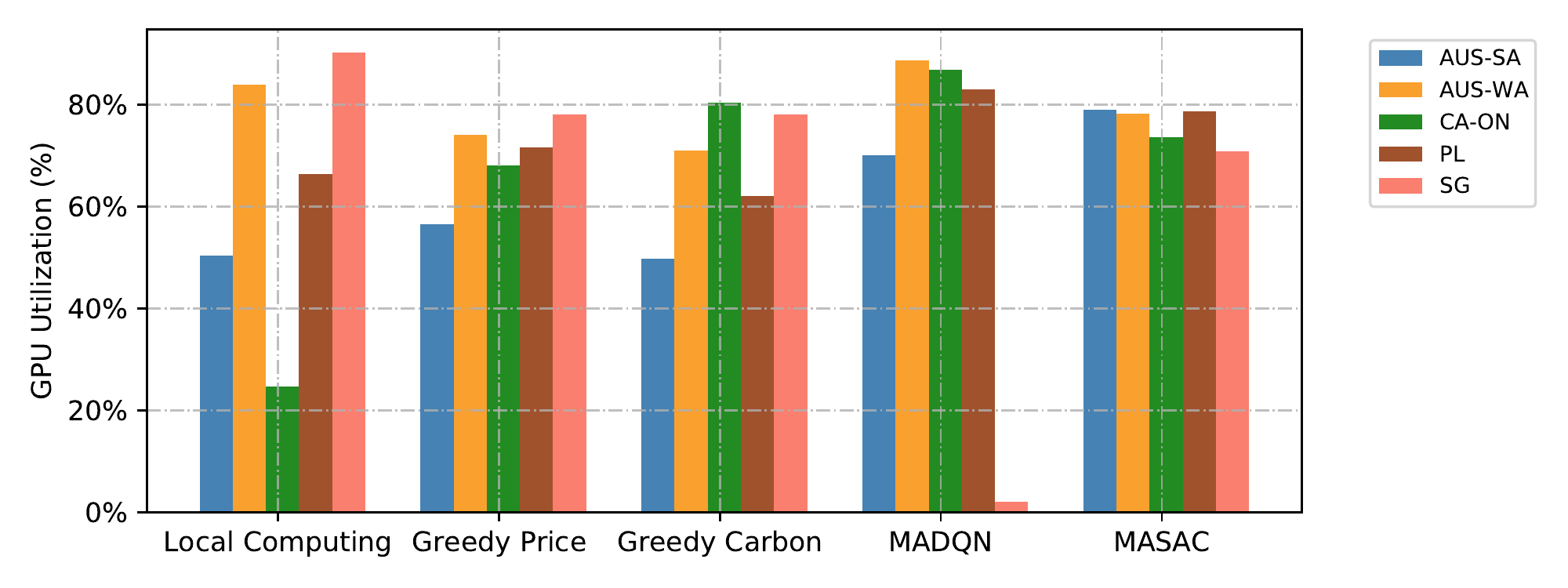}
\end{subfigure}
\caption{Data center GPU utilization of different scenarios.}
\label{fig:u}
\end{figure}

\begin{figure}[t]
%
\centering
\begin{subfigure}{.4\linewidth}
  \centering
  \includegraphics[width=\linewidth]{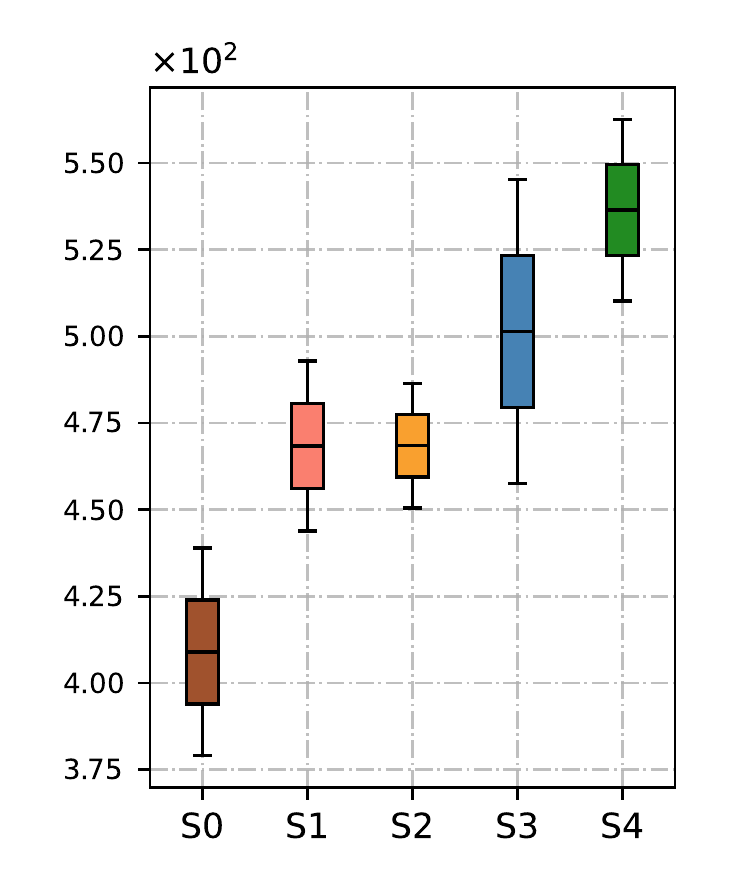}
  \subcaption{Cumulative reward (USD).}
  \label{subfig:cumulative_rew}
\end{subfigure}
\centering
\begin{subfigure}{.4\linewidth}
  \centering
  \includegraphics[width=\linewidth]{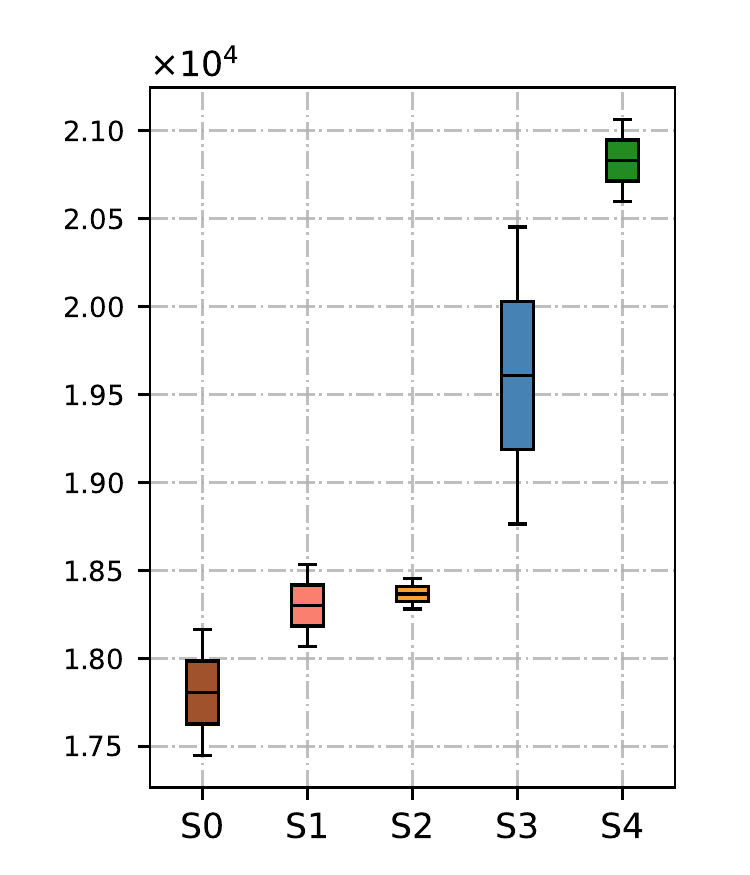}
  \subcaption{GPU hours. \newline \quad}
  \label{subfig:gpuhour}
\end{subfigure}
%
%
\centering
\begin{subfigure}{.4\linewidth}
  \centering
  \includegraphics[width=\linewidth]{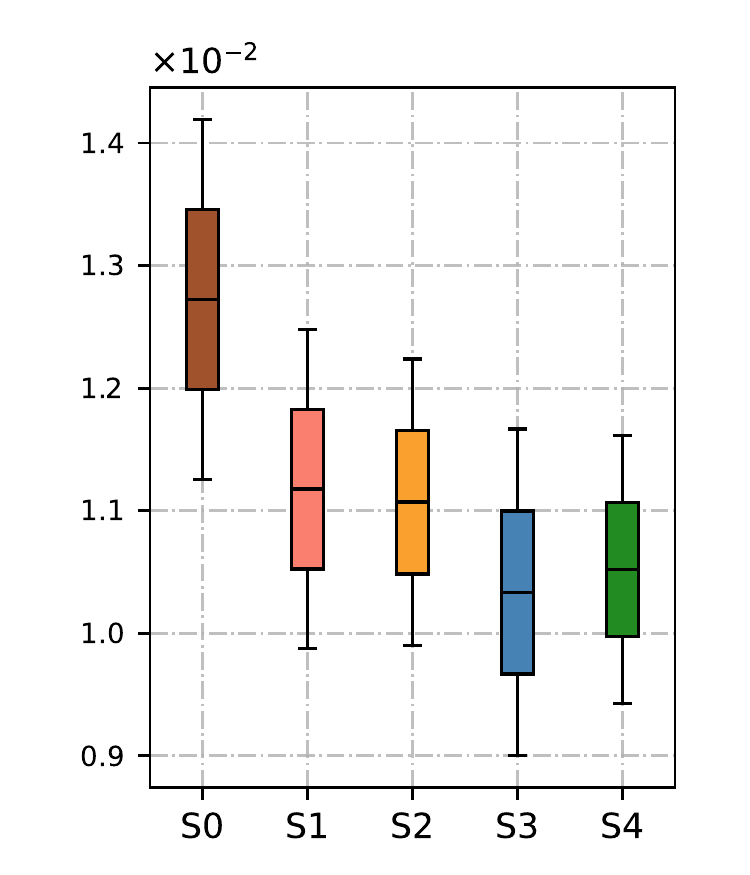}
  \subcaption{Cost efficiency\\(USD per GPU hour).}
  \label{subfig:costpergpuhour}
\end{subfigure}
\centering
\begin{subfigure}{.4\linewidth}
  \centering
  \includegraphics[width=\linewidth]{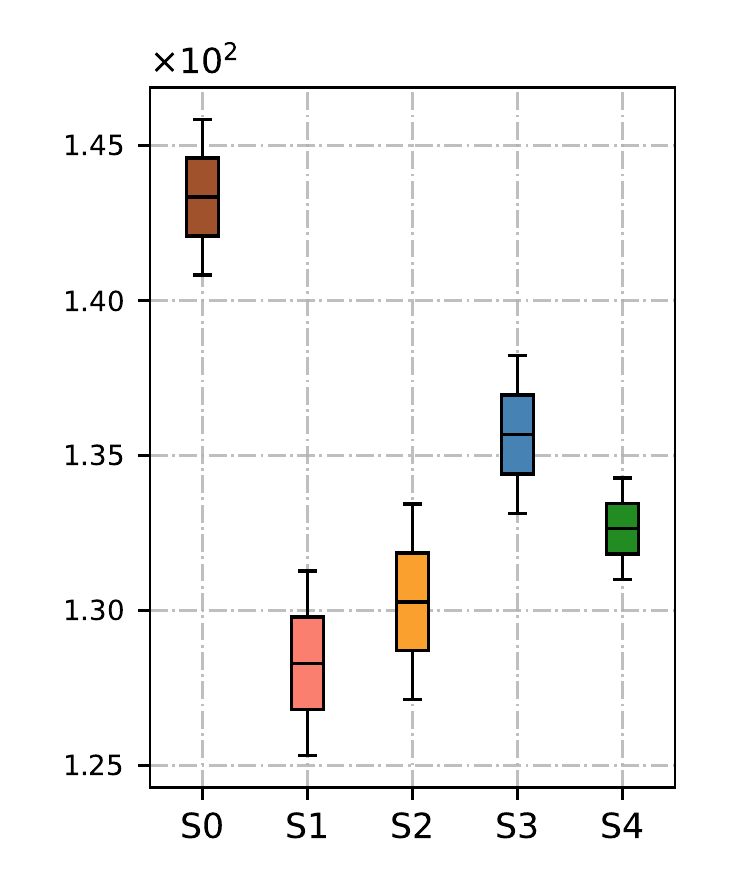}
  \subcaption{Carbon efficiency\\(gCO$_2$ per GPU hour).}
  \label{subfig:carbonpergpuhour}
\end{subfigure}
\caption{Testing performance of different scenarios \\ (S0: Local Computing, S1: Price Greedy, S2: Carbon Greedy,\\ S3: MADQN, S4: MASAC).}
\label{fig:test_performance}
\end{figure}

We illustrate the scenario comparisons during training, validation, and testing in Figs. \ref{fig:train_performance}-\ref{fig:test_performance}. The gradual increase and stabilization of average step reward during training in Fig. \ref{subfig:step_rew} indicate improved policy and convergence, reflecting the effectiveness of training algorithms. MADQN achieves half of the step reward as MASAC, which is mainly because the policy learned by MADQN postpones queuing jobs more frequently, thus leading to a larger number of steps. MASAC and MADQN converge after 10 and 40 epochs in the training. The improvement of learned policies is also visualized in the validation environment as Fig. \ref{subfig:val_rew}. Compared to local computing, all scenarios with job migration have a higher cumulative reward (i.e., system utility), suggesting that the benefits associated with job migration outweigh the costs incurred. MASAC achieves the highest reward (i.e., 28.6\% improvement over local computing) with a rapid and stable learning process. On the contrary, MADQN has a severe performance fluctuation, suggesting that agents may encounter challenges in accurate value estimation.

To understand the scheduling strategies, the GPU utilization during validation is presented in Fig. \ref{fig:u}. Greedy algorithms improve the demand imbalance and utilization of the data center, CA-ON, which demonstrates substantial advantages in energy price and carbon intensity. MADQN offloads most jobs in SG to other locations due to the fact that SG has the highest energy price and relatively high carbon emissions most of the time. Severe resource contention in other data centers is observed, which results in a lot of decision steps for queuing jobs as well as failed jobs. MASAC manages to balance the utilization among data centers and allocate more GPU time.

The algorithms are further tested for different workloads and time periods as Fig. \ref{fig:test_performance}. It can be seen that the high reward of MASAC is mainly attributed to the revenue from GPU time, which is up to 15.6\% higher than other algorithms. The cost and carbon efficiency, denoted by the average energy cost and carbon emissions per GPU hour, are competitive. It implies that MASAC agents wisely select the destination data center considering real-time energy price and carbon intensity. More importantly, the transmission cost of MASAC is decreasing over the training, indicating its ability to avoid unnecessary job migration.

\section{Conclusion}\label{con}
In this paper, we have studied algorithms for scheduling AIGC workloads among geo-distributed data centers. Compared to local computing, it is observed that job migration could bring about significant economic and environmental benefits. The proposed MASAC algorithm has superior performance over others, which balances multiple objectives including maximizing resource utilization, minimizing energy and transmission costs, and reducing carbon emissions.

\bibliographystyle{IEEEtran} 
\bibliography{citation} 

\end{document}